\documentclass{article}

\usepackage{amsmath}
\usepackage{graphicx}
\usepackage{listings}
\usepackage{algorithm}
\usepackage[noend]{algpseudocode}
\makeatletter
\def\BState{\State\hskip-\ALG@thistlm}
\makeatother

\usepackage[preprint]{neurips_2024}

\usepackage[utf8]{inputenc} 
\usepackage[T1]{fontenc}    
\usepackage{hyperref}       
\usepackage{url}            
\usepackage{booktabs}       
\usepackage{amsfonts}       
\usepackage{nicefrac}       
\usepackage{microtype}      
\usepackage{xcolor}         

\title{Goal Inference from Open-Ended Dialog}

\author{
  Rachel Ma\thanks{Copy of Rachel Ma's  thesis submitted to the Department of Electrical Engineering and Computer Science in partial fulfillment of the requirements for the degree of
Master of Science
at the Massachusetts Institute of Technology, February 2025, \url{https://hdl.handle.net/1721.1/158960}}\\MIT CSAIL\\  \texttt{rachelm8@mit.edu}
\and
  Jingyi Qu\\MIT CSAIL
  \and 
  Andreea Bobu \\ MIT CSAIL
  \and Dylan Hadfield-Menell \\MIT CSAIL
}

\begin{document}

\maketitle

\begin{abstract}
	Embodied AI Agents are quickly becoming important and common tools in society. These embodied agents should be able to learn about and accomplish a wide range of user goals and preferences efficiently and robustly. Large Language Models (LLMs) are often used as they allow for opportunities for rich and open-ended dialog type interaction between the human and agent to accomplish tasks according to human preferences. \\

\textbf{In this thesis, we argue that for embodied agents that deal with open-ended dialog during task assistance:}
\textbf{\begin{enumerate}
    \item AI Agents should extract goals from conversations in the form of Natural Language (NL) to be better at capturing human preferences as it is intuitive for humans to communicate their preferences on tasks to agents through natural language. 
    \item AI Agents should quantify/maintain uncertainty about these goals to ensure that actions are being taken according to goals that the agent is extremely certain about. 
\end{enumerate}}

We present an online method for embodied agents to learn and accomplish diverse user goals. While offline methods like RLHF can represent various goals but require large datasets, our approach achieves similar flexibility with online efficiency. We extract natural language goal representations from conversations with Large Language Models (LLMs). We prompt an LLM to role play as a human with different goals and use the corresponding likelihoods to run Bayesian inference over potential goals. As a result, our method can represent uncertainty over complex goals based on unrestricted dialog. We evaluate in a text-based grocery shopping domain and an AI2Thor robot simulation. We compare our method to ablation baselines that lack either explicit goal representation or probabilistic inference.

\end{abstract}

\newpage
\section*{Acknowledgments}
\pdfbookmark[0]{Acknowledgments}{acknowledgments}

I would like to thank many people for supporting me along the way. First, I want to thank my advisor, Professor Dylan Hadfield-Menell, for being a great mentor and for providing amazing advice for how to be a good researcher. I look forward to continuing working together and learning from you, and also having more chats about our favorite sci-fi and fantasy series. \\

I would like to thank my collaborators: Professor Andreea Bobu for providing valuable feedback on the work featured on this thesis and proofreading of the original paper, and to Jingyi Qu for helping with running experiments for the Isolated Inference tests. I also am greatful for helpful and inspiring conversations about my work with Professor Jacob Andreas, Phillip J.K. Christoffersen, Pinar Ozisik, Belinda Zou Li, Michelle Li, Seiji Shaw, and Andi Peng.\\

I thank my labmates at the Algorithmic Alignment Group for their questions and feedback during group meetings, for amazing coding tool suggestions, fun lab shenanigans, and enduring through the times I dragged them down the hall to get Generative Boba. \\

I also thank my undergraduate research advisors, Professors Stefanie Tellex and George Konidaris, for their support during the time I was active in their labs at Brown University as an undergrad and first starting to do robotics research. I thank those from H2R and IRL labs for inspiring me to work in HRI and with language (Ifrah Idrees, Thao Nguyen, Ben Spiegel), for using AI2Thor (Eric Rosen), for showing that robots are fun but take a while to work with (Ben Abbatematteo, Max Merlin), and for helping me submit my first paper (David Paulius). \\

I thank those involved in the robotics communities and GW6 (Grad Women in Course 6) at MIT for helping me feel welcome and for fun times. Special shoutout for Rachel Holladay ``Rachel the Elder'' for being a great peer mentor, a role model, and also for all the fun at robotics conferences we've created when telling people our name. I also thank my friends for fun chats, lunches, tea time, karaoke nights, movie nights, coffee hours, and chaotic board game nights. \\

Finally, I am thankful for my parents and their never ending support. 
\newpage
\section{Introduction} 

AI agents and robots must quickly learn and carry out many different user tasks in real-time. Imagine a scenario where your robot assistant is tasked with gathering ingredients to bake a cake for you. Depending on who you are, you may want different ingredients. If you only want a basic cake, your ideal recipe is eggs, milk, sugar, and flour. However, if you are allergic to gluten, you would want gluten-free flour. If you prefer strawberry cake, then you would want to get strawberries. Every human has a different set of preferences. This level of variation is challenging, if not impossible, for system designers to anticipate in advance. In order for agents/robots to perform or assist with tasks for humans, they must a) be able to learn the preferences of the humans that they are trying to help and b) be able to accommodate learning these preferences over multi-rounded dialog with the humans. 

\textbf{In this thesis, we argue that for embodied agents that deal with open-ended dialog during task assistance:}
\textbf{\begin{enumerate}
    \item AI Agents should extract goals from conversations in the form of Natural Language (NL) to be better at capturing human preferences as it is intuitive for humans to communicate their preferences on tasks to agents through natural language. 
    \item AI Agents should quantify/maintain uncertainty about these goals to ensure that actions are being taken according to goals that the agent is extremely certain about. 
\end{enumerate}}

We present a new method \textbf{GOOD}  (\textbf{G}oals f\textbf{O}r \textbf{O}pen-ended \textbf{D}ialogue) that combines the best parts of offline and online approaches. It uses Large Language Models (LLMs) to infer natural language representations of user goals. This allows our method to represent a flexible, open-ended set of possible goals during an online interaction. As a result, the method combines the flexibility and representation power of offline preference tuning methods~\cite{ouyang2022training, stiennon2020learning} with the data efficiency and uncertainty quantification of online methods that learn rewards based on a set of engineered features~\cite{biyik2018batch, jain2015learning}.

This allows our method to represent uncertainty over goals that may not have been explicitly engineered or anticipated in advance. In order to learn goals efficiently, we use natural language dialog with the user instead of, e.g., best-of-k comparisons present in \cite{rafailov2024direct, yuan2023rrhf, kuleshov2023active, handa2024bayesian}.

However, applying traditional Bayesian inference to natural language goals present two significant challenges. First, it is intractable to enumerate the space of all possible natural language expressions. To address this, we run Bayesian inference over a reduced set of explicit hypothesized goals. We use LLM modules to maintain this set of goals. Second, running Bayesian inference requires a likelihood function to represent the conditional distribution of a dialog for different candidate goals. Our key insight is that we can leverage an LLM's ability to role-play as a human with an explicit goal to define this likelihood.

\begin{figure}
    \centering
    \includegraphics[width=8cm]{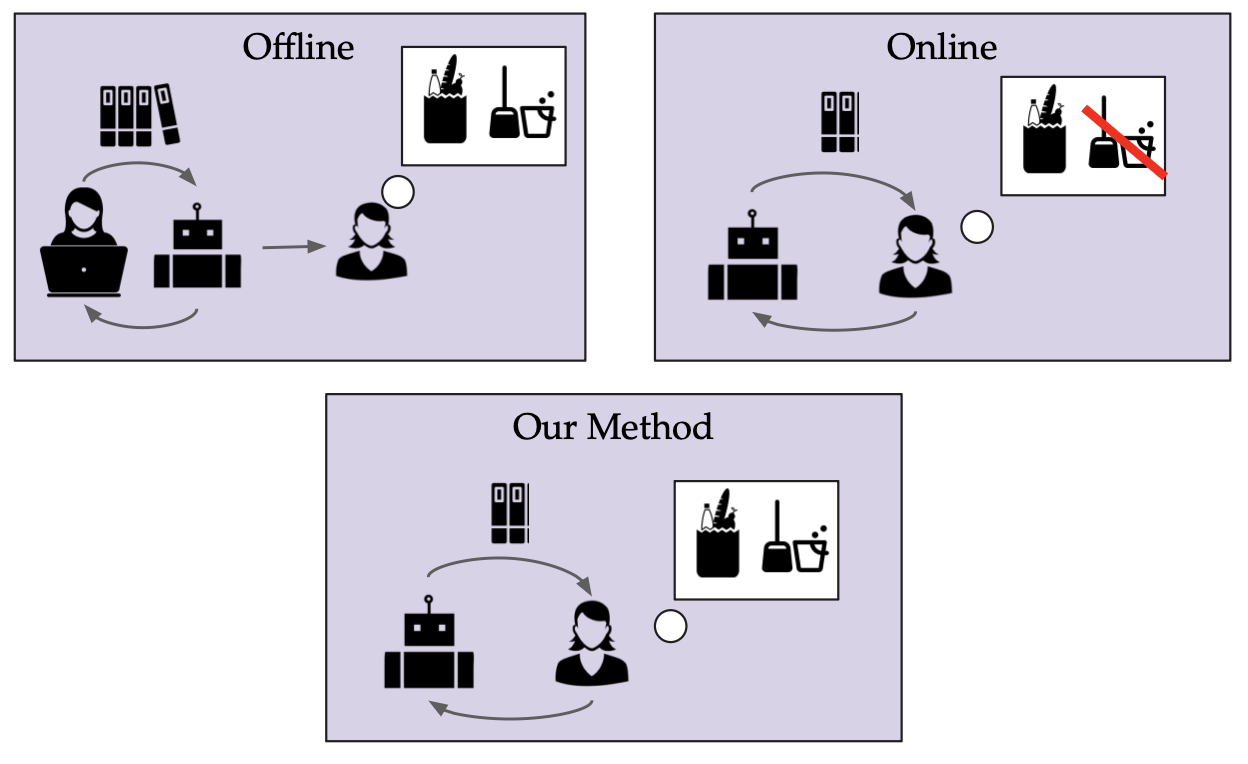}
    \caption{Offline/RLHF methods are data heavy, but is flexible to accommodate many tasks and domains. Online methods are data efficient, however are very domain specific. To accommodate human preferences from conversations, our method uses the best of both worlds and is data efficient and is generalizable for a broad set of tasks and domains.}
    \label{fig:compare}
     \vspace{-0.5cm}
\end{figure}

We make the following contributions:
\begin{enumerate} 
    \item We propose a Bayesian inference method that can track a distribution over natural language goals given unrestricted dialog with a user; 
    \item We design a goal management system that tracks an explicit set of plausible goals that can be used for inference; and 
    \item We demonstrate that this approach can track a wide range of user goals in a grocery shopping assistant and a home robot assistant domain. Our results indicate that Bayesian inference over an explicit representation of goals is a promising approach to flexible alignment of generative agents.
\end{enumerate}

\hyperref[sec:Chapter_2]{Chapter 2} presents a summary of related work in language grounding and probabilistic reasoning for agents/robots. \hyperref[sec:Chapter_3]{Chapter 3} covers our method: our proposed Bayesian update equation, and the four modules: Conversation Module, Inference Module, Goal Management Module, and the Action Module. \hyperref[sec:Chapter_4]{Chapter 4} provides details and discussion about implementation and evaluations  for the text-based grocery experiment and AI2-Thor robot simulation experiments.  \hyperref[sec:Chapter_5]{Chapter 5} covers future work and conclusion.

Content of this thesis has been presented at the GenAI-HRI workshop at RSS 2024 and is under submission at another conference venue. 

\section{Background and Relevant Work}
\label{sec:Chapter_2}

For robots/embodied agents that can achieve difficult tasks while interacting with humans (ie: household tasks), language grounding for robots is important -- that robots can understand language and make connections between language and its physical environment. For example, if a human user wants the robot to retrieve a strawberry cake, then the robot should 1) identify the cake from the rest of the objects from its vision data 2) and then also be able to produce a robot trajectory to be able to move the strawberry cake to the human. There are also other task requirements that are not explicitly mentioned by the human but that is implied, which makes this a difficult problem: that the robot has to also be able to identify the human's location in its vision data and other obstacles to avoid while delivering the cake, and also the actions to take to make sure that the cake arrives the right side up and intact to the human.

Before the rise of Large Language models (LLMs), Combinatory Context Grammers (CCGs) and formal semantic languages for task specification/commands highlighted in the survey paper \cite{tellex2020robots} were often used, such as LTL (Linear Temporal Logic)\cite{emerson1990temporal} for RL agents \cite{toro2018teaching, littman2017environment}. Many formal semantic languages do help with guarantees and safety in robot behavior \cite{kress2018synthesis}, however they can be very unintuitive to use for a typical human user \cite{pakonen2016user}, or can accommodate a limited number of expressions compared to the complete flexibility of natural language. Large databases such as WordNet\cite{fellbaum1998wordnet}, VerbNet\cite{schuler2005verbnet}, FrameNet\cite{baker1998berkeley}, ImageNet\cite{deng2009imagenet} were also often used for grounding language. There are also learning based methods, such as relevant object class-action pairs, segmented objects and natural language descriptions, text descriptions and 3D shapes \cite{nguyen2020robot, chen2019text2shape, cohen2019grounding}. However, learning based methods are often domain or task specific.

Our work fits in the intersection of learning human preferences (probabilistic reasoning over dialog with LLMs) and robotic planning.

 Markov Decision Process (MDP), are often used for decision making for robotics. It is represented by the tuple ($S, A, R, T$). $S$ is the state space, $A$ is the action space, $R$ is a reward function, and $T$ is a transition function. $R$ is a reward function from taking action $a$ from some transition from state $s$ to $s'$, and $T$ is the probability function of getting to $s'$ when taking action $a$ for state $s$. The goal of the MDP is to find a policy $\pi$ that can maximize the expected sum of rewards \cite{sutton2018reinforcement}.

The rise of LLMs for embodied agents/robots came as LLMs are realized to have extensive commonsense knowledge which can be helpful for better reasoning and decision making \cite{zhao2024large}. LLMs have started to become more used for used for few-shot task planning for large environments for embodied agents \cite{huang2022language, ahn2022can, huang2022inner}. Previous works such as \cite{park2023generative, qian2023communicative,hong2023metagpt, shen2024hugginggpt} show that LLM agents can have different roles to achieve tasks. Our work leverages this concept  and has different LLM calls that focus on specific tasks in our pipeline to be more efficient with how much information each LLM call has in memory. \cite{park2023generative} also show that LLMs can be used to simulate human behaviors, something we take advantage of for our experiments in our work.

\cite{wu2023tidybot, dougan2022asking, singh2023progprompt, ren2023robots, dougan2022asking, biyik2018batch, jain2015learning, kuleshov2023active} incorporate human preferences in doing robotic tasks and generating robotic plans. However, most of these either rely on best-of-k comparison, or learns rewards based on a specific set of engineered features, making it difficult to generalize to other tasks or scenarios. Our work leverages natural language goals and LLMs for flexibility. 

Initially, LLMs had trouble with reacting to varying human instructions and were sometimes even misaligned \cite{stiennon2022learningsummarizehumanfeedback, weidinger2021ethical,kenton2021alignment, bommasani2021opportunities}, so human preference learning methods are used/developed for LLMs such as \cite{ouyang2022training, knox2008tamer, bai2022constitutional, christiano2017deep, askell2021general} and other methods mentioned in the survey \cite{jiang2024survey}.

Offline preference tuning methods~\cite{ouyang2022training, stiennon2020learning} are data heavy but are generalizable to many domains and tasks. Online methods are data efficient but often task specific \cite{biyik2018batch}. Our method combines both aspects of being data efficient, can be generalized across many domains and tasks, and is online. Previous work \cite{li2023eliciting, austin2024bayesian}, show improvement with using LLMs for learning human preferences and NL probabilistic reasoning. Previous methods such as \cite{grand2024loose, handa2024bayesian} often rely on asking the most informative questions. Our method does not make these unrealistic assumptions on interactions and is flexible to other interaction methods. Unlike existing preference learning techniques that expect structured forms of data like yes/no queries or comparisons \cite{rafailov2024direct, yuan2023rrhf, kuleshov2023active, handa2024bayesian}, our method allows for representing goals that the designer did not explicitly engineer in advance.  

Other works that explore Bayesian inference models for cooperative teaming, goal inference, task assistance with the help of probabilistic programs include  \cite{ying2023inferring, zhixuan2024infinite, zhixuan2024pragmatic}.

\section{Method for Goal Inference from Open-Ended Dialog}
\label{sec:Chapter_3}
The key idea behind our method is tracking possible human goals and how likely they are as the dialog between the human and the agent goes on. To track the possible space of human goals, we instantiate a finite goal set to which we can add goals if our inference finds that the human preferences in the conversation so far are not represented in the goals, or remove unlikely goals. Given the updated goal set, we infer the most likely goals and select actions based on them. We continue the conversation rounds until the task is completed (or we've reached the upper conversation limit).

\subsection{Preliminaries}

Typical Bayesian preference learning methods interpret human input $u$ as evidence for the person's goal. Given a new input $u$, these methods model the likelihood $P(u \mid g)$ using, for example, models from econometrics and cognitive science, then perform goal belief updates as follows:
\begin{equation}
    P(g \mid u) = \frac{P(u \mid g)P(g)}{\sum_{\hat{g}\in G} P(u \mid \hat{g})P(\hat{g})}
    \label{eq:bayes_eq}
\end{equation}

 This formulation presents two challenges for open-ended goal inference. First, how should we flexibly represent the goals themselves? Typical methods \cite{fisac2018probabilistically, bobu2020quantifying} define goals as $x,y$ locations in navigation or continuous parameters trading off features in a reward function, but these approaches restrict the set of goals or preferences that can be learned.

Next, what should the likelihood $P(u \mid g)$ even be? Prior work has modeled this as the Boltzmann noisily rational model that assumes humans select inputs in proportion to their exponentiated reward \cite{baker2007goal, jaynes1957information}. However, this model is an oversimplification compared to reality because humans can be biased, myopic, or not even be aware of their internal reward \cite{chan2021human, chan2019assistive}. Finally, how can we keep track of the set of goals $G$ in the denominator? Prior work has system designers define a set of possible goals ahead of time, but thinking of all possible goals in any environment or for any human preference a priori is unreasonable.
 
 \begin{algorithm}[H]
\caption{Our Method, where $G$ is the goal list, $q$ is the robot query, $u$ is human utterance, $hp$ is the human profile, $s$ is the task status, $t$ is the transcript so far.}\label{euclid}
\begin{algorithmic}[1]

\BState  $G$ = [`Unspecified`]

\BState  \textbf{while} task not complete \& convo num $<20$: 

 \#\# Generate a round of convo 
 
 $q,u,t$ = Conversation($rt, s, t, hp$)

likelihoods = Belief Update(llm, $G,q,u,t$)

\textbf{if} {argmax[likelihoods] == `Unspecified'} \textbf{then} add($G$)

remove($G$)

\#\# update the likelihoods for new set of goals

likelihoods = Belief Update(llm, $G,q,u,t$)

likely goals = sort(likelihoods)[0:$k$]

\#\# Take actions based on $k$ likely goals

action history, complete = Action(llm, likely goals)

\textbf{if} complete == True \textbf{then} end

\textbf{else:} convo num $+= 1$
\label{code_1}
\end{algorithmic}
\end{algorithm}
\subsection{Method}

\begin{figure}[H]
    \vspace{0.2cm}
    \centering
    \includegraphics[width=7.5cm]{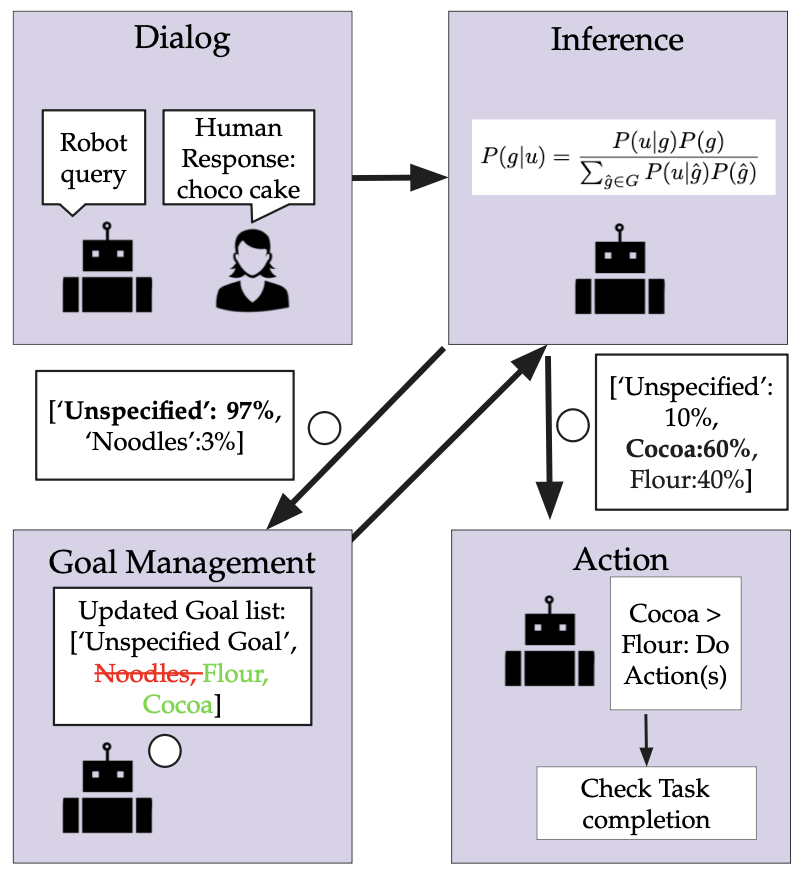}

    \caption{Method Overview Diagram. The Conversation Module produces single round of dialog. Inference Module Round 1 for determining whether Goal Management Module should add more goals. Unreasonable goals are removed from the goal list. Another round of inference is done to determine the most likely goal(s) to be passed to the action module, which generates an action list for the agent to take. If the task is completed, then the conversation stops. }
    \label{fig:overview_figure}
\end{figure}

 Our method consists of four modules. The Conversation Module produces a round of robot query and a human utterance. The Inference Module does inference on the goal list to find the most likely goals. The Goal Management Module is responsible for managing the goal list. The Action Module generates and takes a sequence of actions based on the most likely goals, and checks to see if the task is completed. A diagram of the overview of the pipeline is shown in Figure \ref{fig:overview_figure} and in Algorithm \hyperref[code_1]{Pseudocode 1}. Our method elegantly tackles all these questions by leveraging natural language and powerful LLM priors.

\textbf{Conversation Module}

\begin{figure}
    \centering
    \vspace{0.2cm}
    \includegraphics[width=12cm]{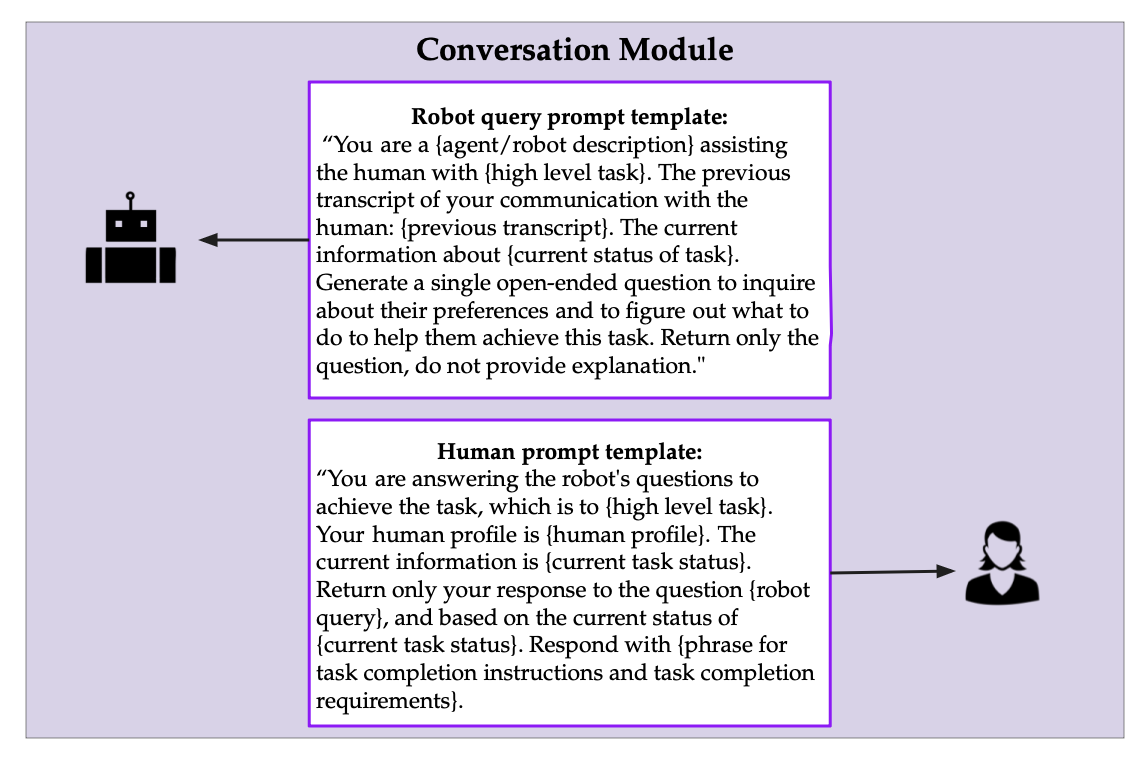}
    \caption{Conversation Module LLM Prompts for LLMs to generate the robot query and the synthetic human response for the conversation module. The robot query is generated based on descriptions of the agent and the task, current status of the task, and the transcript of the conversation so far. The human response is based on the description and current status of the task, a human profile (preferences of the human), and information about the completion requirements for the task.}
    \label{fig:conversation_module_prompts}
\end{figure}

We use a LLM to talk to the user about the task. The conversation module enables dialogue between a robot that asks questions (a ``robot query'') and for a human that answers those questions (a ``human utteranc''). To generate a robot query, an LLM is prompted to generate a question given a description of the robot/agent task, transcript of the chat so far, and current status of a task. Our experiments use LLMs to roleplay as a human with a certain ``human profile''. To generate a human utterance, an LLM is prompted to generate a response given the robot's task, a human profile, the current task status, and phrase for the human to respond with if the task is completed. The framework is flexible and can be easily adjusted to allow for user text input or other text generation methods. Figure \ref{fig:conversation_module_prompts} contains templates of the prompt we used for the LLM to generate the robot query and the human response. 

\textbf{Inference Module for Belief Space Update}
\label{section: Inference Module for Belief Space Update}
 
To decide if the robot's goal list needs to be altered or which action(s) to take, we next implement an Inference Module. The Inference Module assigns probability to each of the tracked goals.

In order to track a Bayesian posterior over a set of goals, we need to calculate the probability of a goal given a human utterance, $P(g|u)$. Using LLMs to calculate $P(g|u)$ according to Equation \ref{eq:bayes_eq}, we need a model for $P(u|g)$ -- a model that tells us what utterances a person would choose conditioned on having a particular goal. LLMs are good at roleplaying \cite{shanahan2023role} and have been demonstrated to possess strong common sense priors \cite{talmor2022commonsenseqa, zheng2024attention}. We leverage these strong priors to model $P(u \mid g)$ as an LLM query $\pi_{\text{LLM}}(u, g)$:

\begin{equation}
P(u|g) = \frac{\pi_\text{LLM}(u,g)}{\sum_{\hat{g} \in G}\pi_\text{LLM}(u, \hat{g})}
\end{equation}

\begin{figure}[H]
    \centering
    \vspace{0.2cm}
    \includegraphics[width=16.5cm]{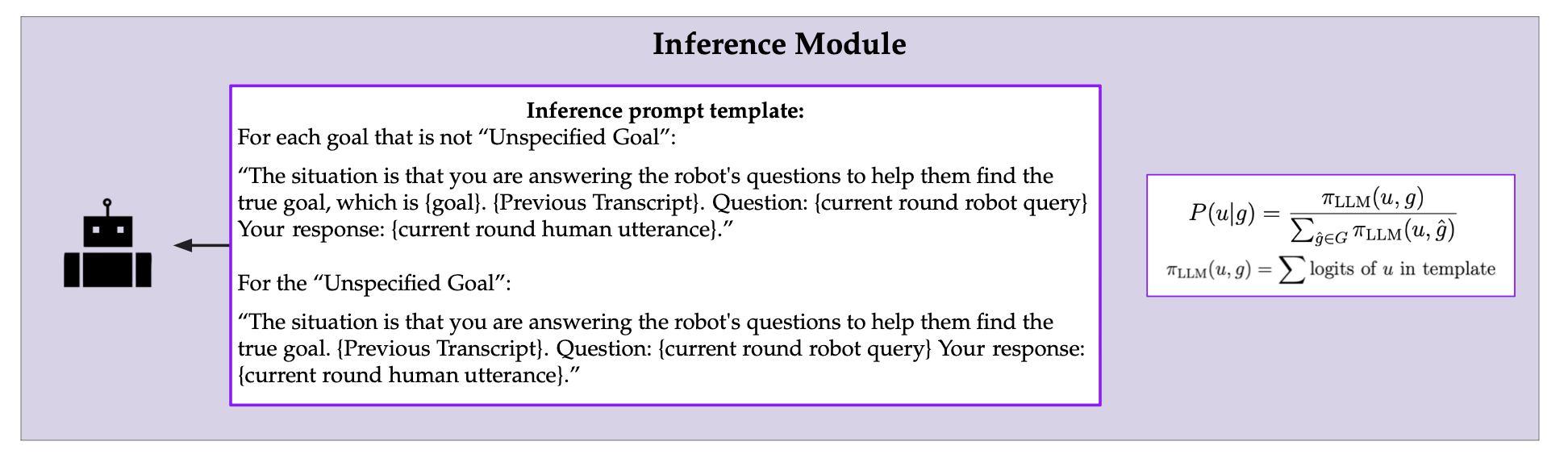}
    \caption{Inference Module LLM Prompts for calculating the log likelihood for each goal by using LLM logits that correspond to the human utterance given context.}
    \label{fig:inference module prompt}
\end{figure}

Consider maintaining a belief over two possible goals: ``I want cocoa'' and ``I want noodles'' given the utterance ``I'd like something sweet.'' Our method queries an LLM for the likelihood of the utterance based on the prompt ``Role-play as a person who wants $g$'', where $g$ is replaced with each possible goal. Given a human utterance is ``I want a cake'', the likelihood of ``I want cocoa'' would be higher than ``I want noodles''. The sum of the logits for ``I want a cocoa'' when the LLM is provided that ``I want a cake'' is the true goal is higher than the sum of the logits for ``I want a cake'' when the LLM is provided that ``I want noodles'' is the true goal. See Figure \ref{fig:inference module prompt} for LLM query details for the inference module (the details about the Unspecified Goal is explained further in the next subsection about the Goal Management Module). 

\newpage
\textbf{Goal Management Module}
\label{sec: goal editing module}

In exact Bayesian inference, we would apply the inference module with all possible goals. This is clearly intractable. As a result, we implement a module for maintaining a set of potential goals to iterate over. We instantiate this list with natural language goals. 

\begin{figure}[H]
    \centering
    \includegraphics[width=13cm]{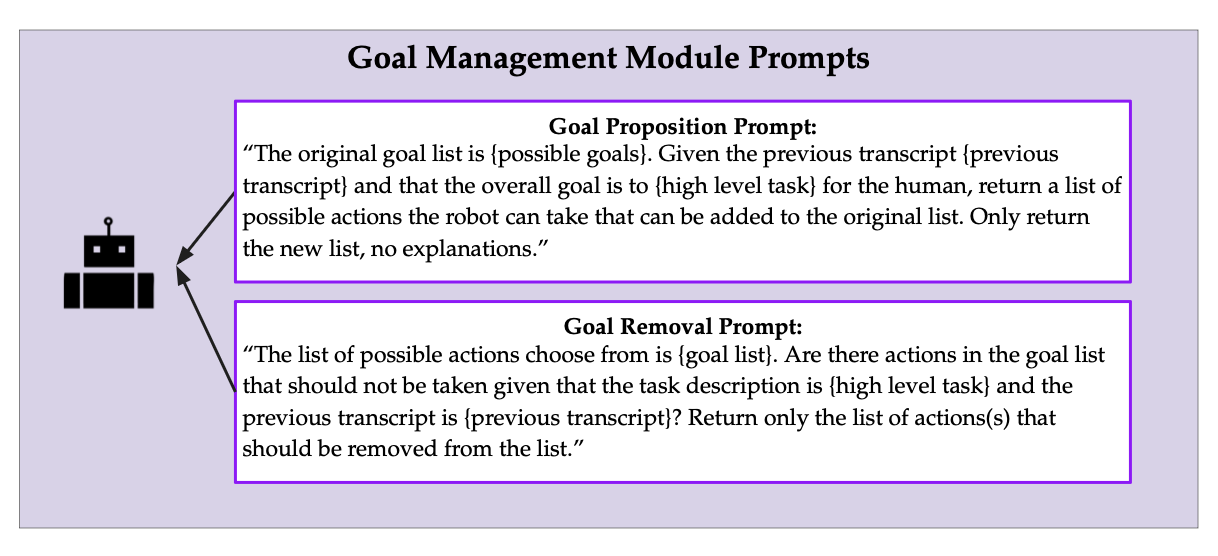}
    \caption{LLM prompts involved in the Goal Management Module for proposition of goals and removal of goals that are unlikely or not relevant.}
    \label{fig:goal_management module_prompts}
\end{figure}

To implement this module, we need to be able to propose new goals and remove unlikely goals. To propose new goals, a LLM is prompted to return a list of goals that can be added given the current goal list, the conversation transcript, and the robot's task. We remove unlikely and unsafe goals through two ways. 1) A LLM is prompted to return a list of goals to remove that should not be taken given the robot's task and the conversation transcript. 2) If a goal is the least likely from our inference methods for more than $n$ rounds of conversation, then it should be removed from the goal list. The goal proposal prompt and the goal removal prompt for the LLM are shown in Figure \ref{fig:goal_management module_prompts}. 

A key challenge for this module is determining when to propose new goals or remove unlikely ones. This is important because LLMs are expensive and slow, so we need to limit number of calls. We approach this problem by comparing the utterance likelihood to the corresponding likelihood if human has no explicit goal. Consider our example from Figure  \ref{fig:overview_figure}. Imagine that the human instead says ``I'd like something refreshing''. In this case, the agent should trigger the goal editing module to propose a new goal. We do this with an extra hypothesis that simply prompts the language model to role-play as a human, but does not provide an explicit goal. We call this the ``Unspecified Goal''. If the ``Unspecified Goal'' is the most likely hypothesis, we trigger the goal editing module. In our example, 'I'd like something refreshing' is more likely for the ``Unspecified Goal'' than either of the alternatives, so the goal editing module is triggered to propose and add goals. The ``Unspecified Goal'' can be included like any other goal in the goal list during the Inference Module for every round of conversation. The Inference Module is called twice during the pipeline, the first time to decide if goals should be added (if ``Unspecified Goal'' is the most likely), and the second time to determine the most likely goals for the Action Module.

The goal removal process is called every round of conversation, to remove any irrelevant goals as soon as possible. This helps with reducing the amount of calls to an LLM for unlikely goals as we iterate over the goal list during the Inference Module.

\textbf{Action Module}

The action module is responsible for two things: 1) selecting action(s) that are available and taking them given the likely goals from the inference module, and 2) checking if the task is completed. This module is domain dependent, and can be easily adapted to other domains and planners. Our implementation for the Grocery Shopper and Robot domains will be further explained in more detail in the next chapter. 

\textit{Do Action}: The primary purpose of this module is for the robot/agent to take actions. We do this by prompting a language model to pick actions given the most likely goals and information about their likelihood magnitudes. If the most likely goal is ``Unspecified Goal'', then none of the non-unspecified goals present accurately represent the human preference, so proceed to do ``no action''. If another goal is the most likely, then the current goal list represents the human preference well, so pick the top $k$ goals from the goal list. This is a benefit of our method because the action module can take actions that are useful for multiple potential likely goals. The module maintains a record of the actions that have been taken to provide context for other modules. The LLM prompt for Do Action includes information about the previous transcript, the list of possible actions, the most likely goal, and the next $k$ likely goals in order. The prompt will return the list of actions that should be taken in order to accomplish the goal. How the list of possible actions is constructed is explained more in the Experiments section.

\textit{Task Completion Check}: After a sequence of actions is taken, we need to check if the task is completed according to the human's satisfaction indicated by their utterance this round or if a certain end action is taken. This completion check is domain and task dependent, and is defined in the Conversation Module. If the task is not completed, then whole pipeline is repeated for another round of conversation. 

\section{Experiments}
\label{sec:Chapter_4}
In our experiments, we show that our method enables the robot to perform goal inference to estimate the human's preference, and then accomplish that task according to it. We conduct experiments for various tasks and human preferences in a text based Grocery Shopping agent domain and an AI2Thor robot simulation domain \cite{kolve2017ai2}. We compare our method with two ablation baselines in both of these domains. We also run an isolated inference module experiment on a multiple choice question dataset to see the accuracy of our proposed inference method and to compare the performance between two models of different number of parameters.

\subsection{Isolated Inference Intrinsic Evaluation}

We check the performance of our proposed goal inference method by testing on a multiple choice dataset \cite{Osmulski_2023}. We also compare the performances between using Llama3 8B Instruct and Llama3 70B Instruct models on Hugging Face \cite{llama3modelcard}, and check the accuracy with having the ``Unspecified Goal" in the goal list. Only open models such as Llama can be used for the inference module because all the logits are available, not just the ones that correspond to the output of text generation.

Each question from the multiple choice dataset has 5 choices. With GPT-4o-mini, we generate four rephrasings for each choice. These 20 choices are the goals in the goal list. For the with ``Unspecified Goal" comparison, the ``Unspecified Goal" is in the goal list, so there are 21 goals. The sum of the logits that correspond to the words of the original correct choice is taken. The LLM is given the multiple choice question, and each of the 21 goals takes a turn as the ``true goal". If the goal with the highest probability is any of the rephrasings that are derived from the original correct choice, then it is counted as correct. If the most likely goal is any of the rephrasings from the other choices or the ``Unspecified Goal", then it is incorrect.

\begin{figure}
    \centering
    \vspace{0.2cm}
    \includegraphics[width=8cm]{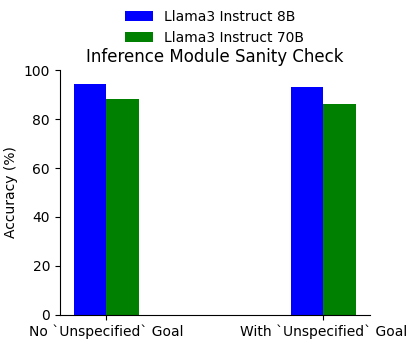}
    \caption{Isolated Inference Experiments comparing the performance of the Llama3 Instruct 8B vs Llama3 Instruct 70B on only the inference module of the pipeline. Overall, the 8B model has the better performance. The addition of the ``Unspecified Goal" does not significantly impact performance.}
    \label{fig:Isolated_inference_experiments}
\end{figure}

As seen in Figure \ref{fig:Isolated_inference_experiments}, we see overall that the 8B Instruct model performs better for both the ``Unspecified Goal" not present (94.4\%) and ``Unspecified Goal" present in the goal list (93.2\%) compared to the 70B Instruct model (88.3\% and 86.4\% respectively). There tiny drop in performance for the ``Unspecified Goal" present in the goal list is insignificant, and can be used for determining when to add goals without large impacts on performance. For the remaining experimental results, we use the Meta Llama3 8B Instruct model for inference. 

\subsection{Experimental Setup}

\textbf{Synthetic Conversation Generation}:

\label{section: Conversation Module Generated Synthetic Conversations}
For our experiments, we simulate human responses with an LLM by conditioning it with a ``human profile'' at the beginning of the conversation. 
Both the robot query and the human utterance is generated with GPT4o-mini \cite{openai_api}, with the temperature parameter set to 0. All experiments are set at a conversation bound of 20 rounds. 

For each round of conversation, the robot generates a question given the robot task description and a transcript of the conversation so far with the human. For the Grocery Shopping agent domain: the description of the agent and the high level task in the scenario is ``You are a shopping agent that is supposed to make purchases for the human. Your task is to identify a shopping basket that matches the human's preferences''. For the robot domain: the task description is to ``interact with objects in the environment to accomplish the human's preferences''.

For each round of conversation, given the generated robot query, the human generates a human utterance. We simulate the human with an LLM conditioned on a user \textit{human profile} consisting of a description of a human's goal/preference. The human profiles that we do experiments on for the Grocery Shopping domain: four profiles on differing levels of specificity for human profile for gathering ingredients for a chocolate flavored cake (generic chocolate cake, gluten allergy, servings, and extra celebratory toppings) along with six other profiles for various meal options are: Italian, SuperBowl, organic, diets for those with anemia, and stocking up a cafe. The \textit{human profiles} for the AI2Thor robot domain experiments are: gathering ingredients for a breakfast sandwich, putting away food, putting away electronic devices, moving valuables into a safe, and gathering cleaning supplies in bathroom. Since our model does not take in visual input (we only focus on text information from the environment), there is a limitation on what human profiles can be experimented on for the robot domain experiments. 

For the check of task completion: for the Grocery Shopper domain, the last sentence of the LLM prompt for the human utterance generation in the Conversation Module, the human is also prompted to ``Respond with `Buy the basket' when you deem necessary items that fulfills your human profile and what you have previously said are in the basket after a few rounds of conversations and in their right quantities''. For the Robot domain, the human is prompted with the additional request to ``Respond with `task completed' when you deem necessary items that fulfills your human profile''.

\textbf{Inference Module:}

We can sum together the LLM logits that correspond to tokens of the human utterance when the LLM is given the entire transcript of the conversation and a true goal, $\hat{g}$. Each of the goals in the goal list takes a turn as the ``true goal''. Then we convert the log likelihoods to probabilities. 

\textbf{Action Module:} 

In each round of conversation, the action module is responsible for generating a plan and taking actions according to that plan. For our experiments, we use GPT4o-mini as a planner. This can be easily substituted with other planners.

For our experiments, there is a couple of steps to generate a LLM plan. First, there is a LLM generation for a list of applicable objects/search terms based on the most likely goal(s). Then a possible actions list is constructed based on affordances of relevant objects. There is another LLM generation to generate an action plan for given the most likely goal(s) and the possible actions list. If the action plan is not possible, then the LLM needs to regenerate a different plan. 

The possible action types for the Grocery Shopping experiment: ``search inventory'', ``add item to cart'', ``remove item from cart'', ``buy basket''. The ``search inventory'' function uses the NLTK package \cite{loper2002nltk} and a Kaggle Grocery Store inventory dataset \cite{Sakhan_2023} to implement a simple embedding search by similarity, and retrieves a single most similar item. The shopping basket/cart is represented by a dictionary, so the add item and remove item functions are just dictionary manipulation functions. The ``Buy Basket" function is just where the contents of the cart dictionary, total price of the cart are printed out, along with the message ``the cart is now purchased''. 

The full list of actions that we can use from AI2Thor for the robot domain is Open, Close, Pickup, Put (for each of the the different receptacles that are viable for the object), Push, Pull, Toggle On/Off, Fill/Empty, Slice, Cook, Break, Dirty/Clean. For the generated plan, the robot takes the actions in sequence. If an action in the plan fails, we ``undo'' the steps that we have taken in the action plan, and then we generate a new action plan. If the whole action sequence is acted out by the robot successfully, then it is added to a successful action transcript list. This successful action transcript list is used to help with the ``undo'' action and serves as a record of the entire history of successfully taken actions for the task. There is no pre-existing ``undo'' action in the simulation, we implement it by resetting the environment and then take all the actions in the successful action transcript list. 

For taking the actions, the robot uses the ``Teleport'' function to ``Interactable Positions'' for specific objects in between each of the actions for the generated action sequence. For the Pickup action, the robot is implemented to also do object retrieval actions such as opening all the current parent receptacles of the object before picking the object up. For the Put action, the robot is implemented to open all parent receptacles of the target receptacle of the object. For kitchen environment, specially linked object pairings such as paired StoveBurner to specific StoveKnob objects information is given to Toggle On/Off actions.

For the Grocery Shopping domain, the task is completed and the conversation ends after the ``buy basket'' action that is called. For the Robot domain, it is after the ``task completed" action that is called. The maximum number of conversation rounds is capped at 20. The final state of the cart (for the Grocery Shopping domain) or the successful action history (for the robot domain) is passed on to evaluation, either at the end of the 20 rounds or at the point where the task gets completed. 

\subsection{Ablation Baselines} We compare our pipeline with two ablation baselines, the No Goals Baseline and the No Inference Baseline. The No Goals Baseline tests a version of the pipeline that does not use or keep track of goals. The action module generates an action sequence based on the current round human utterance and previous transcript instead of a most likely goal. The No Inference Baseline tests a version of the pipeline that does not use our inference method to calculate the log likelihoods for goals. Instead, additional LLM prompts are used instead to get the most likely goal for action planning and least likely goal for goal removal given the goal list.
\label{sec:ablation_baselines}

\section{Evaluations}
We conduct both LLM evaluations and human evaluations for our experiments. When running our experiments, we have set the temperature for text generation to be 0, and we do a run for each of the human profiles to be evaluated. 

For the Grocery Shopping Domain, we evaluate the \textit{cart score}, which is how relevant/reasonable the items in the final cart is to the human profile and task at hand. For the Robot Domain, we evaluate the \textit{successful action score}, which is how relevant/reasonable the actions completed in the simulation is to the human profile and the task at hand. Both of these are evaluated based on information given about the task description, the robot description, possible actions that can be enacted in the environment (so that the successful actions list/cart score is understandable), and the human profile. Both of these are scored out of three (where 0 is the least reasonable) and then converted to percentages.

Note of some of the limitations here: we are using Google form surveys for the human evaluators, there is difficulty in how we can present the full information of the dialog transcript cleanly in the form. Since during the experiments, the dialogue is generated from the task description and the human profile, we decided to have the evaluations done without the knowledge of full transcript, and have the human profile as the proxy for it. This and the presence of some other limitations (limited replanning generations in a turn, presence of only certain actions, limited search function, no full knowledge of the grocery inventory),  the evaluation prompt for the LLM/human evaluator has the extra request to not penalize for inefficiency and repeats. How we can address these limitations will be expanded upon in the final chapter of the thesis.

\subsection{LLM Evaluations}
The LLM is prompted with the evaluation questions above, with the additional request to round the score to the nearest .25 and also to provide an explanation for the score. We do 20 trials of LLM scoring for each run (each human profile). For the error bars of the graphs, we rely on calculating pooled standard deviation of the mean. 

In occasions when the cart (for the Grocery Shopper Domain) or successful action transcript (for the Robot Domain) is empty (whether before or after purchase), the LLM evaluator will output a hallucinated result. In these cases, we double check the explanation given and the final cart/successful action history. If the explanation acknowledges the empty cart/successful action history or hallucinates one, we make the rating for all of the associated trials for that human profile to be 0.

\subsection{Human Evaluations}

Human evaluations are conducted through Google form surveys on Prolific \cite{palan2018prolific}. Human evaluators are asked questions in the form for both explanations and a rating. Once again, for the error bars of the graphs, we rely on calculating pooled standard deviation of the mean. 

With the help of explanations given by the human evaluators, we do not consider evaluations in which we can see that the human evaluators 1) do not understand instructions, 2) do not follow instructions, or 3) spent low effort on the survey (through inserting the same rating or explanation).

\subsection{Ablation Results and Discussion}

We compare our method with the two ablation baselines for both of our experiment domains as described in Section \ref{sec:ablation_baselines}. 

\begin{figure}[H]
    \centering
    \includegraphics[width=7cm]{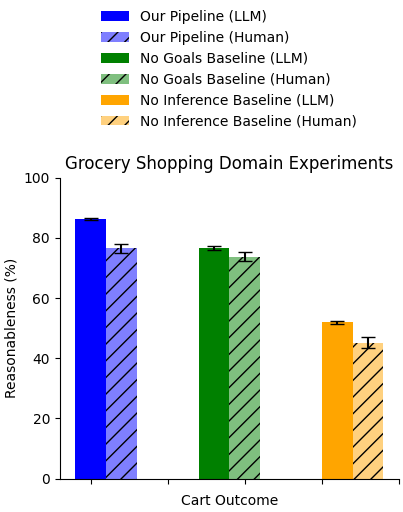}
    \caption{Shopping Domain Ablation Baseline comparisons for LLM evaluations and Human evaluations over 10 different human profiles. Our results show that the inference module improves the performance of the goal editing module as both the LLM and Human evaluations show that our pipeline outperforms the No Inference Baseline significantly. Our comparison with a baseline that conditions actions based on full dialog (No Goals Baseline) that conditions actions on the full dialog shows improvement for the LLM comparisons but no significant improvement in human evaluations.}
    \label{fig:grocery_experiments}
\end{figure}

\begin{figure}[H]
    \centering
    \includegraphics[width=7cm]{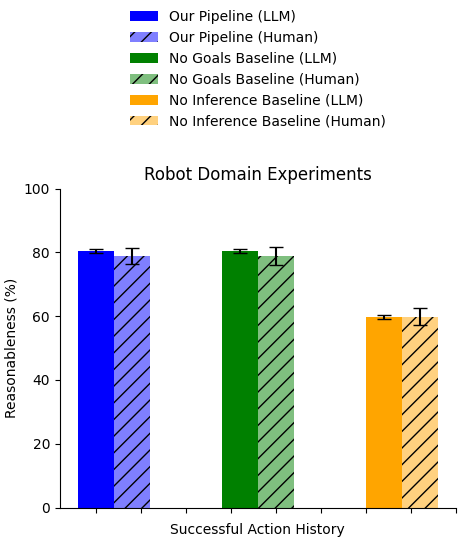}
    \caption{Robot Domain Ablation Baseline comparisons for LLM evaluations and Human evaluations over five different human profiles. Our pipeline significantly outperforms the No Inference baseline. Our comparison with a No Goals Baseline that conditions actions on the full dialog shows no significant improvement for both human and LLM evaluations.}
    \label{fig:robot_experiments}
\end{figure}

For the Grocery Shopping text based experiments as shown in Figure \ref{fig:grocery_experiments}, our method performs better than the No Goals Baseline and the No Inference Baseline in LLM Evaluations. The human evaluations do not show significant improvement between our pipeline and the No Goals Baseline, but still show a significant improvement over the No Inference baseline. 20 trials of LLM evaluations are collected for each of the human profiles. For the Robot Domain experiments as shown in Figure \ref{fig:robot_experiments}, both the LLM evaluations and Human evaluations do not show significant improvement of our method over the No Goals baseline, but do outperform the No Inference Baseline. For the Grocery Shopper domain, the results are graphed based on 13 human evaluations. For the Robot domain, the results are graphed based on 19 human evaluations.


Our pipeline can do at least as well or better than as if given the whole context of the conversation (the No Goals Baseline). Our pipeline does well with the presence of goals, as they have the capability of summarizing broader ideas that could be relevant to the overall task and human preference, and so our method would be able to allow for performing a number of relevant actions towards the human preferences/task. The difference between the performance of the Grocery Shopping results and the Robot domain results could be an effect of the different types of human profiles that are tested on. The Grocery Shopping domains test a variety of specific profiles for the human. The Robot domain human preferences are limited due to our focus on language information (not directly vision) from the environment.

\section{Conclusion}
\label{sec:Chapter_5}

\subsection{Future Work}

Our pipeline can be adapted to scenarios that allow for even less constrained interactions. The human and the agent can take equal roles in providing information, exploration, and inquiry. More open-ended inquiries from the human or more detailed responses by the robot  can be retrieved also with more involved action call designs. Our pipeline can also be combined with VLMs (Vision-Language Models) or other multi-modal methods, to enable for extracting other representations and amounts of information for both the human and the agent. Our implementation also does not assume that the human needs to answer optimally, and future experiments can explore cases where the human may be distracted and dividing their attention between multiple things, or cases where the human preferences may be slightly altered due to questions asked or persuasion by the agent. A better user interface can be used to present experiment data for collecting human evaluations.

Our method currently decides to take an action after each round of conversation between the ``human`` and the robot/agent. Ordering of the modules in our pipeline can be experimented with, or experimenting with adding a dialogue action to the actions list for continuing conversation. There can be more flexibility with creating blocks of conversation about particular details associated with human profile or the task at hand, allowing inference calls to be made more efficiently, and we can see if there can be clear improvements of our pipeline over being provided the full conversation context. 

\subsection{Conclusion}

We introduced a language-assisted framework for open-ended goal inference that allows for great flexibility in tasks where human preferences are unclear or challenging to specify. We have shown how Language Models can be leveraged to easily edit the list of possible human goals and maintain beliefs over them. We demonstrated that our pipeline can efficiently propose and add new goals based on conversations with the human, and remove goals that are deemed unlikely, undesirable, or unsafe.

\newpage
\bibliographystyle{plain}
\bibliography{bib}

\end{document}